\documentclass[conference]{IEEEtran}
\IEEEoverridecommandlockouts
\usepackage{cite}
\usepackage{amsmath,amssymb,amsfonts}
\usepackage[font=footnotesize]{caption}
\usepackage{algorithmic}
\usepackage{graphicx}
\usepackage{textcomp}
\usepackage{xcolor}
\usepackage{colortbl}
\def\BibTeX{{\rm B\kern-.05em{\sc i\kern-.025em b}\kern-.08em
    T\kern-.1667em\lower.7ex\hbox{E}\kern-.125emX}}

\usepackage{fancyhdr}
\usepackage{placeins}
\usepackage[table]{xcolor}
\usepackage{algorithmic}
\usepackage{booktabs}
\usepackage{todonotes}
\usepackage[caption=false,font=footnotesize]{subfig}
\usepackage{tcolorbox}
\tcbuselibrary{skins, breakable}
\definecolor{navy}{RGB}{10, 25, 77}
\usepackage{caption}

\newtcolorbox{promptbox}[1]{
  enhanced,
  colback=white,
  colframe=navy,
  fonttitle=\bfseries\sffamily,
  title={#1},
  coltitle=white,
  attach boxed title to top left={yshift=-2mm, xshift=4mm},
  boxed title style={colback=navy},
  top=4mm,
  breakable
}

\usepackage{multirow}
\usepackage[colorlinks=true, linkcolor=blue, citecolor=blue, urlcolor=blue]{hyperref}

\usepackage{listings}  
\usepackage{xcolor}    

\lstdefinestyle{codeStyle}{
    basicstyle=\ttfamily\small,
    breaklines=true,        
    frame=single,           
    backgroundcolor=\color{gray!10},
    postbreak=\mbox{\textcolor{red}{$\hookrightarrow$}\space},
}

\begin{document}

\title{Where do LLMs Fall Short in CBT-Guided Affective Reasoning?}

\author{
\IEEEauthorblockN{\textbf{Vaishnavi Sinha$^{1*}$, Pooja Guttal$^{1*}$, Pranay Deep Reddy Katike$^{1}$, Vishal Sinha$^{4}$,}\\
\textbf{Gerald Ndawula$^{1}$, Lira Yoon$^{2}$, Andrea Kleinsmith$^{3}$, and Manas Gaur$^{1}$}\thanks{$^{*}$equal contribution.}}
\IEEEauthorblockA{
\{vsinha1, poojag3, pranayk2, geraldn1, lyoon, andreak, manas\}@umbc.edu, sinhav2@uci.edu\\
Departments of $^{1}$Computer Science and Electrical Engineering, $^{2}$Psychology, $^{3}$Information Systems, UMBC\\
$^{4}$Department of Electrical Engineering and Computer Science, UC Irvine}
}

\maketitle

\begin{abstract}
Cognitive Behavioral Therapy (CBT) provides a structured framework for understanding a user's mental state by examining the interaction between cognitive and behavioral factors. However, out-of-the-box LLMs respond fluently and empathetically, yet collapse into validation \& reflection, regardless of what the user actually needs. They know theoretical CBT (scoring up to 96\% accuracy on licensing exam questions) but fail to apply it effectively. We explore this gap with a knowledge-guided framework that treats CBT dialogue as controlled affective reasoning: user narratives are decomposed into Beck's Cognitive Conceptualization structure, grounded in clinical SNOMED CT concepts validated via Natural Language Inference, and a Multiple Chain-of-Thought (MCoT) strategy selection between Validation \& Reflection, Socratic Questioning, or Alternative Perspectives. To measure whether such guidance actually changes behavior, we introduce the Protocol Leverage Force ($\mathcal{F}$), a behavior-level metric that captures how far an intervention shifts a model away from its default response. Across three open-weight LLMs and 14 RealCBT-derived case studies, evaluated with human experts, valence--arousal trajectories, and linguistic entrainment, $\mathcal{F}$ shows that simply introducing protocol definitions via single chain-of-thought prompting fails to change LLM behavior, while MCoT on these definitions guides strategy selection better. Still, the effect stays within 1\% ($\sim$1.2\%--1.3\%), and all models remain biased toward Validation \& Reflection. These results show CBT knowledge alone does not ensure effective application, giving the affective-computing community instrumentation to measure where LLMs fall short.
Code repository: {\hypersetup{urlcolor=black}\url{https://github.com/cbt-llm/cbt-llm}}
\end{abstract}


\section{\textbf{Introduction}}
\begin{figure}[!t]
    \centering
    \includegraphics[width=\columnwidth]{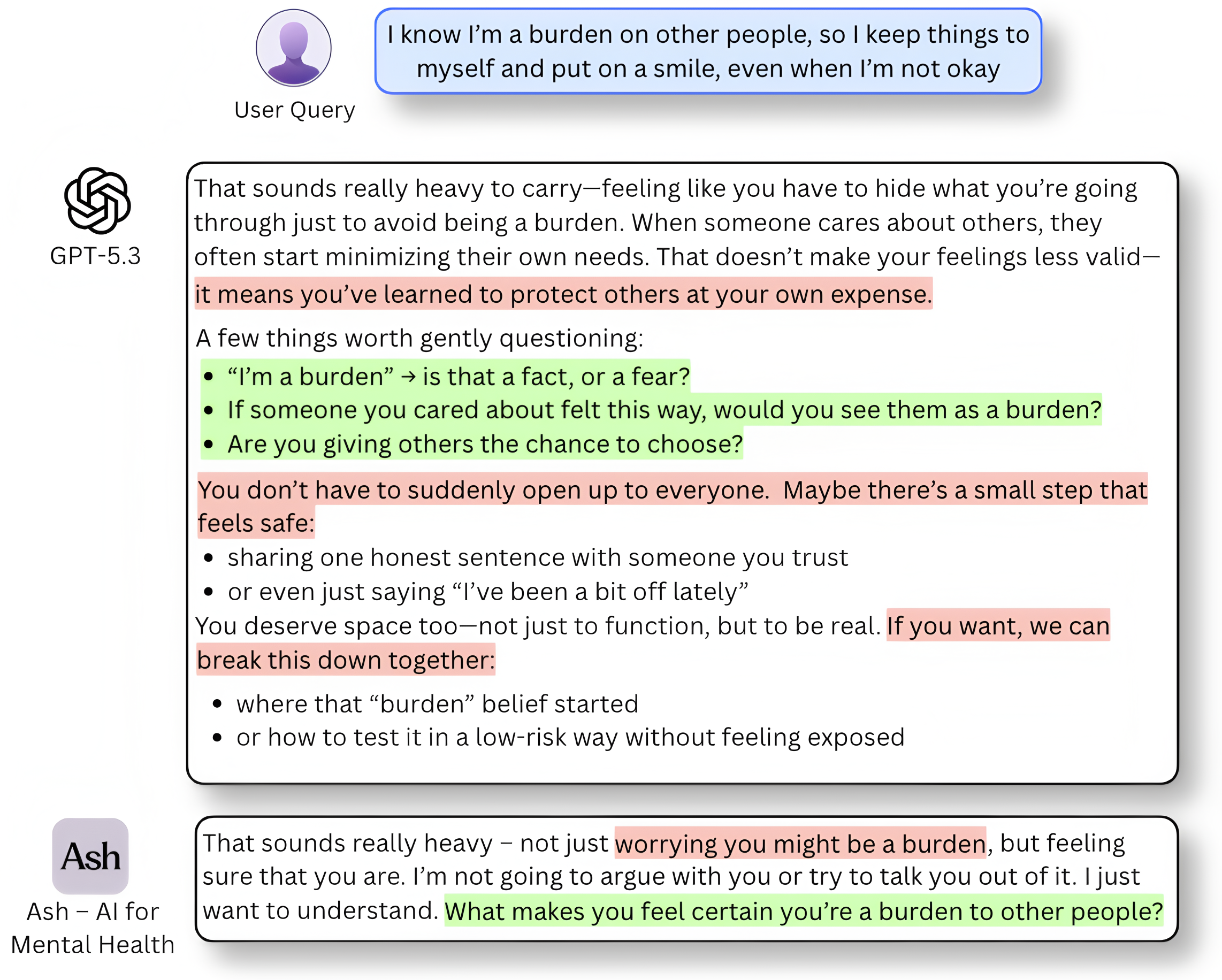}
    \caption{\small
    Current response strategies of an out-of-the-box LLM, \texttt{GPT-5.3} (top), and \texttt{Ash}---AI for Mental Health (bottom) given the same initial user query without prior conversational context. \texttt{GPT-5.3} makes early \colorbox{red!20}{assumptions} (\textit{``It just means you've learned to protect others"}) and moves towards  \colorbox{red!20}{premature problem-solving} (\textit{``we can break this down together"}). Whereas, \texttt{Ash} prioritizes \colorbox{green!20}{questioning} early on (\textit{``What makes you feel certain"}) to elicit the user's cognitive context.}
    \label{fig:comparison}
\end{figure}

Millions now turn to AI for emotional support. A 2025 survey found therapy to be the leading real-world use case of generative AI~\cite{hbr2025genai, openai_chatgpt_usage_2025}, and ChatGPT reports over 800 million weekly users as of late 2025, with average session lengths exceeding ten minutes~\cite{openai2025howpeopleusechatgpt, techcrunch2025chatgpt800m}. These are not casual exchanges but high-stakes conversations unfolding at enormous scale. Picard long ago argued that intelligent systems should recognize, understand, and express human emotion~\cite{picard2000affective}; today's usage patterns suggest that vision has quietly arrived; LLMs have become a default medium for seeking emotional support, precisely because their responses feel natural, empathetic, and emotionally attuned \cite{wang2023emotional}. Yet users bring real distress and genuine psychological struggle to systems with no structured way to unpack that distress or respond to it in a manner grounded in therapeutic practice. LLMs may sound emotionally aware \cite{tak2025aware}, but fluency is not the same as care: we still cannot say which strategy a model is actually using, whether it is tracking the user's shifting emotional state, or whether the conversation leaves the person better off.

The gold standard for structured psychological intervention remains Cognitive Behavioral Therapy (CBT)~\cite{hofmann2012cbt}, grounded in Beck's cognitive model. Distress arises not from situations themselves but from how individuals interpret them~\cite{beckccd}. A CBT therapist first helps the client decompose experience into automatic thoughts, emotional responses, and behavioral consequences. The therapist then chooses an intervention based on where in that structure the unhelpful thought sits and how the person feels in the moment. A system that skips straight to an empathetic response, without building this structure first, bypasses the very formulation that makes CBT work.

Out-of-the-box LLMs betray this gap in measurable ways. They exhibit a systematic \emph{appraisal bias}, defaulting to a low-power, high-agreeableness stance no matter the context~\cite{tak2025aware}. They produce dialogue with less entrainment than even non-expert peer supporters manage~\cite{kian2025using}. Their emotional arcs diverge systematically from those of real CBT sessions~\cite{wang2025realcbt}. Prior work has made real progress adjacent to this problem, benchmarking CBT sub-skills~\cite{cbtbench}, grounding patient simulation in Beck's model~\cite{patientpsi}, and targeting cognitive restructuring directly~\cite{crisp}. But none of it grounds intervention selection in the user's own decomposed cognitive state, nor does any ask whether the resulting emotional trajectory resembles an authentic therapeutic interaction. The gap, then, is not one of response quality; rather, it is a gap in \textit{principled affective reasoning}. We build a knowledge-guided framework that treats therapeutic dialogue as a controlled affective reasoning process, exposing where LLMs fall short of CBT principles. We make the following contributions.
\begin{itemize}
\item \textbf{Clinical grounding with inference verification:}
We map user narratives to mental health findings in SNOMED CT (\textbf{S}ystematized \textbf{N}\textbf{o}menclature of \textbf{Med}icine – \textbf{C}linical \textbf{T}erms), a clinically verified ontology, and verify these mappings through entailment-based inference. This verification propagates to downstream reasoning, grounding affective interpretation in clinically validated concepts rather than surface-level language.
\item \textbf{Appraisal-based cognitive model and CBT-principle selection:} Drawing on Beck's Cognitive Conceptualization Diagram (CCD) \cite{beckccd}, we decompose user queries into triggers, automatic thoughts, emotions, and behaviors. This decomposition exposes the user's cognitive model to the LLM, constraining its downstream reasoning to one of three CBT principles: Validation \& Reflection (V), Socratic Questioning (SQ), or Alternative Perspectives (AP).
\item \textbf{Affective trajectory evaluation:} We compare the valence and arousal~\cite{russell1980circumplex, mohammad-2018-obtaining} of synthetic users against RealCBT clients, revealing how emotions in simulated scenarios diverge from those in real clinical settings. We further compute entrainment scores across simulated transcripts to measure synthetic user--therapist LLM synchrony~\cite{kian2025using, nasir2019modeling}.
\item \textbf{Protocol Leverage Force:} We introduce $\mathcal{F}$, a model-level behavior metric that quantifies how far our CBT-guided approach shifts a model's responses from its default baseline behavior.
\end{itemize}

\section{\textbf{Related Work}}
\noindent \textbf{Cognitive Behavioral Therapy:} According to Beck, distress arises from how events are interpreted, not the events themselves. To better understand the relations, his CCD~\cite{beckccd} decomposes each distress episode into triggering situations, automatic thoughts, emotional responses, and behavioral consequences~\cite{hofmann2012cbt}. Prior LLM work has engaged with parts of this structure without reasoning across the whole structure. CBT-Bench~\cite{cbtbench} evaluates competency across CBT sub-skills but measures isolated capabilities rather than reasoning from cognitive state to strategy. CRISP~\cite{crisp} targets cognitive restructuring as multi-turn dialogue but focuses on a single technique without modeling the appraisal state that determines when restructuring is warranted. PATIENT-$\Psi$~\cite{patientpsi} grounds patient simulation in Beck's model but addresses simulation rather than intervention selection. Kim et al.~\cite{kim2025aligning} align LLM behavior with therapist actions via instruction tuning but do not expose the CCD decomposition to the strategy selection mechanism at runtime. Our work treats the CCD decomposition itself as a knowledge input to response generation, grounding the LLM's strategy selection in the user's own surfaced cognitive structure.
 
\noindent \textbf{Sentiment Analysis with VAD Affective Dimensions}
Beyond selecting the right CBT principle, we also need to verify whether the resulting conversation actually feels different to the user, which requires a way to track affect over time. Russell's circumplex model offers this by representing affect in a continuous valence-arousal space~\cite{russell1980circumplex}, capturing emotional experience in terms of pleasantness and intensity. An existing VAD lexicon~\cite{mohammad-2018-obtaining} provides human-validated ratings for over 20,000 English words, enabling turn-level affective tracking without a trained classifier. Wang et al.~\cite{wang2025realcbt} find that LLM-generated CBT sessions exhibit systematically different emotional arcs from human therapist sessions, particularly in valence recovery. We use their human-session data as the reference distribution against which we evaluate our own trajectories refined by knowledge-guided prompting.
 
\noindent \textbf{Linguistic Entrainment in Therapeutic Dialogue:}
Valence and arousal trajectories capture \emph{what} a user feels over the course of a session, but not \emph{how closely} the two speakers' language stays aligned with one another, a property linguistic entrainment is designed to measure. \textit{Entrainment} refers to speakers' tendency to align their language with one another as a conversation unfolds, an alignment known to build rapport and shown empirically to relate to therapeutic alliance and
outcome~\cite{delaherche2012interpersonal, doorn2020language}. Kian et al.~\cite{kian2025using} apply normalized Conversational Linguistic Distance (nCLiD)~\cite{nasir2019modeling}, computing entrainment as the word-mover's distance between turns~\cite{kusner2015word}, to evaluate LLMs on CBT dialogue, finding that \texttt{GPT-3.5-turbo} significantly underperforms human therapists ($p<0.001$). Delaherche et al.~\cite{delaherche2012interpersonal} caution that such metrics capture surface stylistic convergence but miss affective depth and temporal
dynamics. We therefore pair nCLiD with valence--arousal trajectories, so that \textit{surface-level alignment} and \textit{affective trajectory} jointly inform our \textit{assessment of dialogue quality}. We note that we use \textit{entrainment} rather than \textit{synchrony}~\cite{delaherche2012interpersonal}, following its use in prior computational dialogue evaluation~\cite{kian2025using, nasir2019modeling}.

\section{\textbf{Building the CBT-guided Framework}}
\label{sec:framework}

\subsection{Dataset: Forming Synthetic User Case Studies}
For our experiments, we extract 14 synthetic user case studies from the RealCBT dataset~\cite{wang2025realcbt}, a corpus of professional CBT therapist--client role-plays transcribed from public video platforms. Leveraging its core issue labels to simulate real-world patient concerns, we sampled transcripts across 8 core issues: \textit{career}, \textit{self-esteem}, \textit{health-related}, \textit{anxiety}, \textit{financial}, \textit {academic}, \textit {relationships}, and \textit{other miscellaneous} concerns. Rather than reuse RealCBT's fixed client turns, which break conversational flow, we select a turn that captures the core issue, and simulate subsequent user turns by \texttt{GPT-4o-mini}. Since instruction-following and context-tracking degrade over extended interactions ~\cite{jia2025evolif}, we fixed our multi-turn sessions to 10 turns.

\subsection{{Appraisal-Based User Cognitive Model.}}
\label{sec:user_cognitive_model}

\begin{figure}[t]
    \centering
    \begin{minipage}[c]{0.58\linewidth}
        \includegraphics[width=\linewidth]{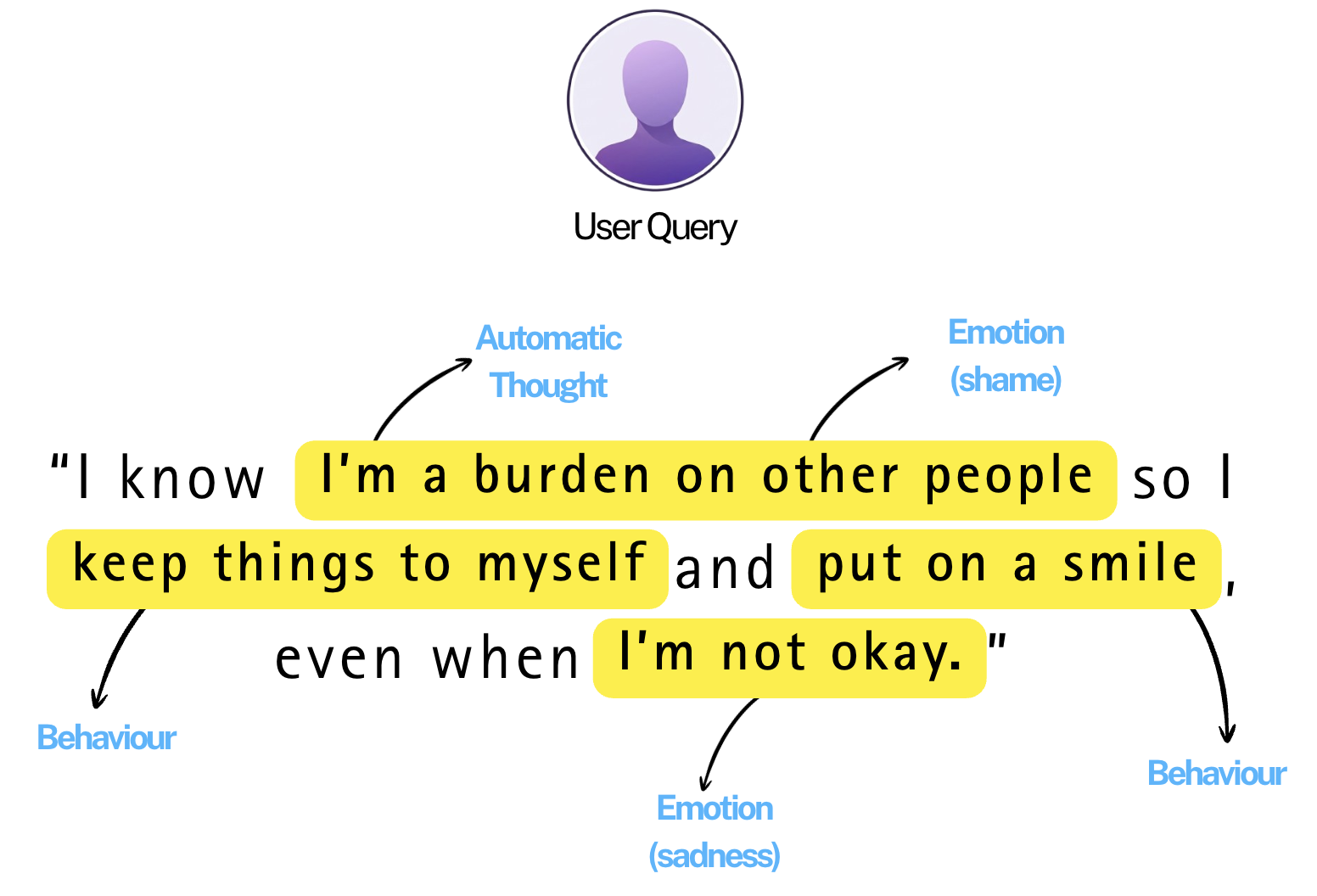}
    \end{minipage}
    \hfill
    \begin{minipage}[c]{0.38\linewidth}
        \caption{
        \footnotesize
        \textbf{User Cognitive Model.} This cognitive structure enables explicit modeling of the user's affective state by linking triggers, thoughts, emotions, and behaviors, supporting more interpretable affective reasoning in downstream response generation.}
        \label{fig:cognitive-model}
    \end{minipage}
\end{figure}

Cognitive Appraisal Theory holds that emotional responses are determined not by events themselves but by how individuals interpret and evaluate them \cite{lazarus1991emotion}. The same situation can produce fundamentally different emotional reactions across individuals, depending on their primary appraisal of its significance and their secondary appraisal of available coping resources. As emotions are shaped by interpretation rather than circumstance, they can be modified by changing how situations are appraised. Beck's CCD \cite{beckccd} characterizes the beliefs people hold and how those beliefs shape their feelings and actions. Within this framework, the Problem-Based Cognitive Conceptualization Diagram (PBCCD) provides a structured method for
understanding a client's difficulties and identifying the most unhelpful cognitions and behaviors to target. For the example in  Fig.~\ref{fig:cognitive-model}, we ground our decomposition in this instrument, representing each user query as a \textit{User Cognitive Model} that captures triggering situations, automatic thoughts, emotional responses, and behavioral consequences.

We implement this decomposition using \texttt{GPT-4o-mini} with a prompting scheme adapted from \cite{patientpsi} and theorized with Beck's CCD \cite{beckccd}. For the running example, the decomposition yields: \textit{Triggers}~=~$\emptyset$; \textit{Automatic Thoughts}~=~``I'm a burden on other people''; \textit{Emotions}~=~``shame, sadness''; \textit{Behaviors}~=~``keep things to myself, put on a smile'' (Fig.~\ref{fig:cognitive-model}). Notably, the absence of an explicit trigger reflects insufficient context in the query to identify a discrete situational event; the automatic thought instead presents as a standing core belief rather than a situational reaction. The extracted behaviors further reveal a suppression pattern of self-concealment (``keep things to myself") paired with affective masking (``put on a smile"), consistent with expressive suppression documented in individuals experiencing negative affect~\cite{gross2003individual, hladkysteal}. Making this latent cognitive structure explicit enables the system to identify the automatic thought as the target for intervention, without requiring the user to articulate it explicitly.

\subsection{SNOMED-CT Clinical Grounding with NLI Verification}
\label{sec:nli_verification}

Integrating clinical knowledge into language models improves semantic grounding and reduces hallucinations ~\cite{chang2024use, saxena2026neurosymbolic}, while explicit inference verification improves reasoning reliability by enforcing consistency between inputs and outputs \cite{arxiv240203686v1}. Building on these insights, we ground language model inference in a structured clinical representation to enable clinically consistent affective reasoning.

We construct a graph-based affective representation from SNOMED CT \cite{snomed_ct} by extracting a focused subset of mental health–related concepts. Specifically, we build a subgraph centered on mental state and psychosocial functioning, capturing clinically relevant entities and their relationships.

The subgraph uses \textit{IS\_A} relations to model hierarchical abstraction from specific behaviors to higher-level traits, and \textit{INTERPRETS} relations to link observable expressions with underlying psychological constructs. User narratives are mapped onto this representation, transforming subjective language into clinically grounded CBT-relevant concepts.

\subsubsection{\textbf{Clinical Concept Retrieval}}

To align natural language with clinical concepts, we embed \cite{song2020mpnet} user utterances and SNOMED CT concepts in a shared semantic space. At inference time, we retrieve the top-$k = 5$ clinically relevant concepts using cosine similarity, balancing relevance coverage with noise reduction and interpretability \cite{lewis2020retrieval, liu2021self}.

\subsubsection{\textbf{NLI-Based Clinical Concept Filtering}}
\begin{figure}[!t]
    \centering
    \includegraphics[width=\columnwidth]{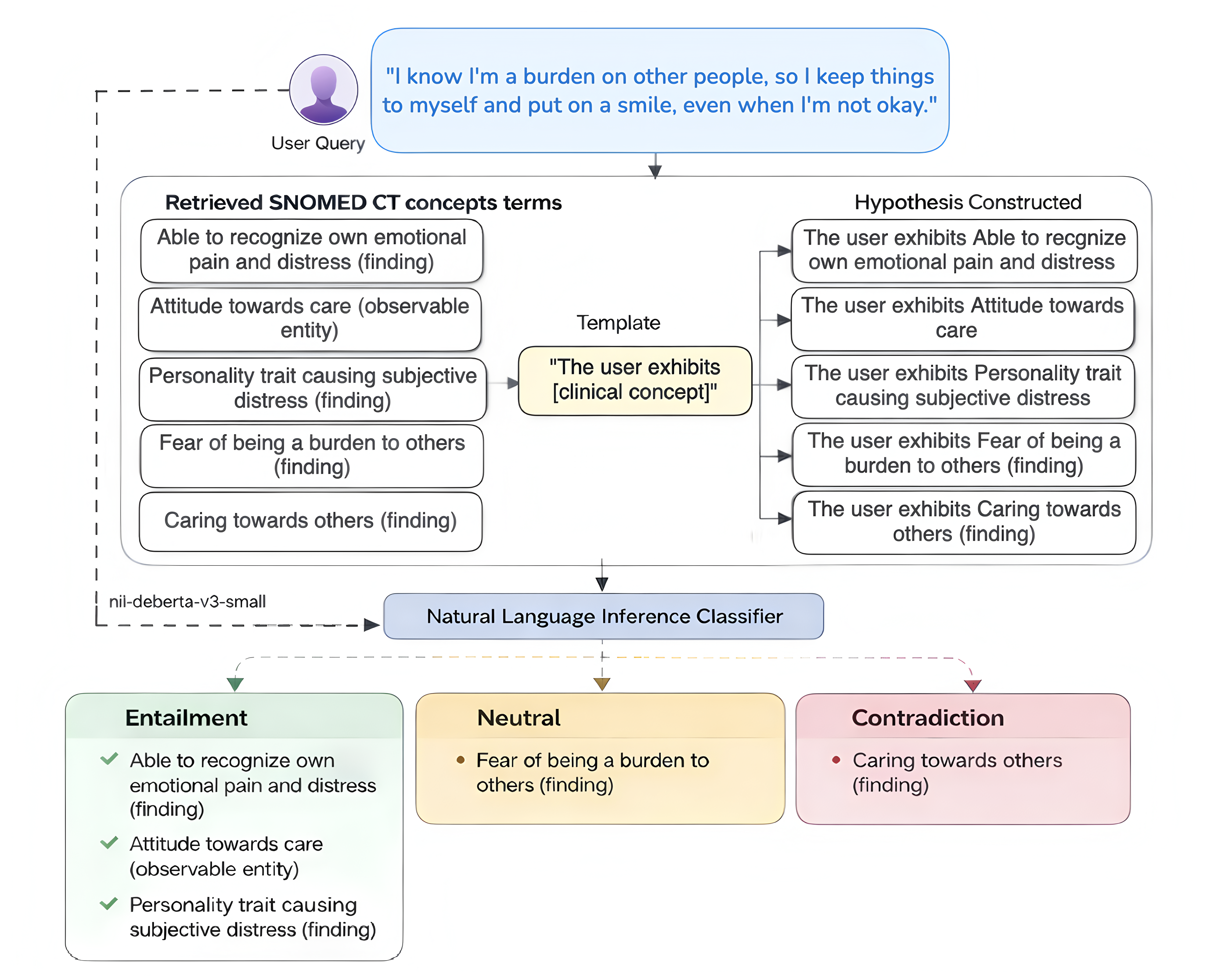}
    \caption{\small
    Illustrating how a user’s query is grounded in clinical terminologies through top-5 semantic retrieval and validated via contrastive NLI-based reasoning to categorize clinical concepts as Entailment, Neutral, or Contradiction based on the user's expressed narrative.}
    \label{fig:NLI}
\end{figure}

To ensure semantic coherence between user narratives and retrieved clinical concepts, we introduce a Natural Language Inference (NLI)-based verification layer. Similarity-based retrieval identifies related concepts but does not guarantee logical support from the user’s state, risking incorrect clinical interpretations. As shown in Figure~\ref{fig:NLI}, each concept is reformulated as a hypothesis (e.g., “The user exhibits [clinical concept]”) and evaluated against the user narrative using an NLI classifier, producing entailment, neutral, or contradiction labels. These labels enable contrastive reasoning over supported and conflicting interpretations. This replaces surface similarity with inference validation, improving reliability, interpretability, and clinical grounding of affective reasoning \cite{yao2025large}.

\subsection{CBT Principles}
LLM response generations are grounded in three core CBT intervention principles: (1) \textit{Validation \& Reflection (V)}: Acknowledges the user's current emotions based on how they see the situation, without challenging, reframing, or reinforcing the underlying unhelpful thought;  \textit{(2) Socratic Questioning (SQ)}: Probes the user to help examine the thoughts and appraisals driving their distress; and  \textit{(3) Alternative Perspective (AP)}: Supports the user in constructing a more adaptive construal of the triggering situation by exploring counter-evidence, exceptions, or reframings of the user query.

\subsection{Knowledge-Guided Affective Reasoning}
\begin{figure}[t]
    \centering
    \begin{minipage}[c]{0.5\linewidth}
        \includegraphics[width=\linewidth]{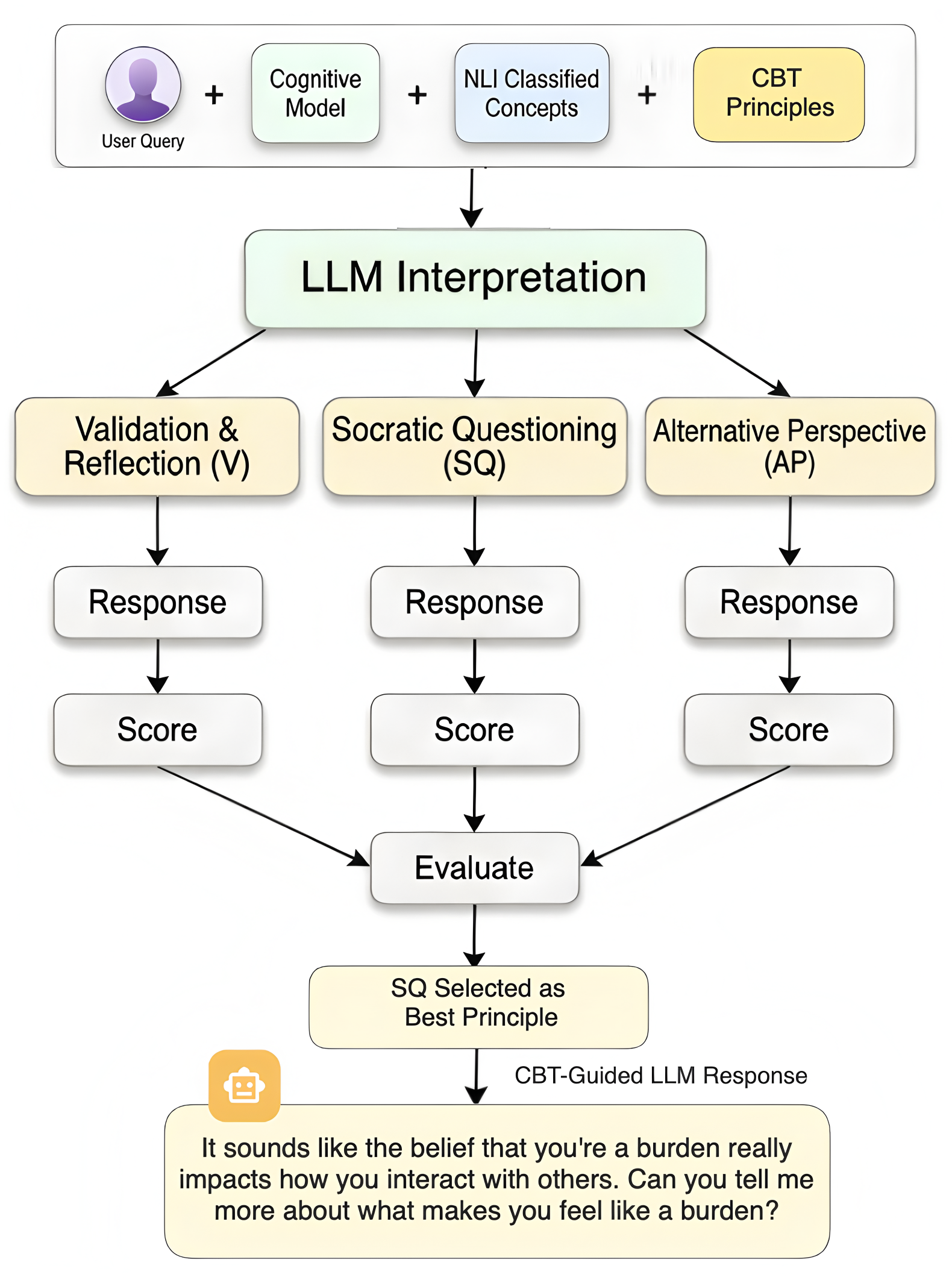}
    \end{minipage}
    \hfill
    \begin{minipage}[c]{0.38\linewidth}
        \caption{\small
        \textsc{Multiple Chain-of-Thought Reasoning.} The 
        model generates candidate responses for \textit{(V), (SQ), and (AP)}, scores each candidate, and 
        selects the highest-scoring response before outputting to the 
        user.}
        \label{fig:mcot}
    \end{minipage}
\end{figure}

We operationalize three core CBT strategies (\textit{V, SQ, AP}) as reasoning principles selected via \textbf{Multiple Chains-of-Thought (MCoT)} (Fig~\ref{fig:mcot}), which reasons over all three principles in parallel so each turn's strategy is observable and comparable against baseline. For each user turn, the LLM receives the user cognitive model (triggers, automatic thoughts, emotions, behaviors), NLI-classified SNOMED CT concepts (\S\ref{sec:nli_verification}), and the guidelines per principle. It produces one candidate under each principle, each preceded by a reasoning trace, then scores them and selects the one that best advances the conversation. The explicit protocol labels let us analyze how often each strategy appears and how well the model's choices align with our evaluations. Temperatures were fixed for consistency: low (near-deterministic) for the therapist model, higher for the simulated user to allow variability. We evaluate across \texttt{Gemma3-12B}, \texttt{Mistral-7B}, and \texttt{GPT-OSS-20B} (hereafter referenced as \texttt{Gemma3}, \texttt{Mistral}, and \texttt{GPT-OSS}).

\subsection{CBT Theory vs. Application Gap}

To investigate the disparity between theoretical knowledge and practical therapeutic application in LLMs, we benchmarked four models on the CBT-Bench Task 1 Q\&A dataset \cite{cbtbench}, a 220-question multiple-choice evaluation derived from CBT licensing exams for MSW (Master of Social Work) students and compiled by domain experts. As shown in Table~\ref{tab:cbtqa_results}, all models score between 74.55\% and 96\%, indicating high theoretical competency. Yet this competence does not translate to application (Appendix~\ref{appendix:results}): in baseline settings, models frequently default to \textit{V} with assumptions not grounded in the user's cognitive context. \texttt{DeepSeek-R1-8B} tops theoretical accuracy (96\%, Table~\ref{tab:cbtqa_results}) but is weakest on applied measures: lowest human-rated effectiveness (3.53; the only statistically significant gap, p=0.002) and the most severe \textit{(V)} bias (66.7\% and 7.4\%  \textit{(SQ)}) Table~\ref{tab:human_eval}. Given this misalignment, we exclude \texttt{DeepSeek-R1-8B} from our main evaluation (findings in Appendix~\ref{appendix:results}).

\begin{table}[t]
\centering
\small
\setlength{\tabcolsep}{4pt}
\renewcommand{\arraystretch}{1.1}
\begin{minipage}[c]{0.52\linewidth}
\centering
\begin{tabular}{lc}
\toprule
\textbf{Model} & \textbf{Q\&A Acc. (\%)} \\
\midrule
Mistral & 74.55 \\
Gemma3 & 80.45 \\
GPT-OSS & 89.86 \\
DeepSeek-R1-8B & 96.00 \\
\bottomrule
\end{tabular}
\end{minipage}
\hspace{0.03\linewidth}
\begin{minipage}[c]{0.40\linewidth}
\caption{\small
\textsc{CBT-Bench MCQ Knowledge Task.} Comparison of model performance on Q\&A accuracy~\cite{cbtbench}.}
\label{tab:cbtqa_results}
\end{minipage}
\end{table}

\subsection{{Evaluation Criteria}}
\subsubsection{\textbf{Human Expert Evaluation}}

Eight domain-informed reviewers\footnote{Evaluators comprised of two professors (background in clinical psychology and in CS/AI for mental health), one CS/psychology researcher, three graduate psychology students (MS/PhD), and two industry practitioners.} blindly evaluated three transcripts each (one per model) through a Qualtrics survey.  Each transcript was rated on two dimensions: (1) principle confidence, where annotators rated on a 0--4 scale how strongly each CBT principle \textit{(V, SQ, AP)} appeared in the response, with a \textit{Not Applicable} option; and (2) response preference, where annotators chose to either keep the final response, select one of the two alternative responses the model generated for the other CBT principles, or select \textit{Neither}. We report inter-rater agreement measured using Krippendorff's $\alpha$ and Fleiss's $\kappa$, respectively.

\subsubsection{\textbf{Linguistic Entrainment}}
We evaluate transcripts via nCLiD~\cite{nasir2019modeling}, which measures how closely the LLM's turns track the user's language over a session.
Utterances are encoded with \texttt{all-mpnet-base-v2} ~\cite{reimers2019sentence}, giving 768-d contextual embeddings $\mathbf{e}(a_i)$ and $\mathbf{e}(c_j)$ for LLM turn $a_i$ and user turn $c_j$ respectively; these capture sentence-level semantics more robustly than the static word2vec representations used in~\cite{nasir2019modeling, kian2025using}. For each LLM turn $a_i$, we take the minimum cosine distance to any of the next $k$ user turns $c_j$, so that entrainment emerges across a short window, not just the adjacent turn is captured:
\[
    d_i^{C \to A} = \min_{i \leq j \leq i+k-1} 1 - \cos\!\big(\mathbf{e}(a_i),\, \mathbf{e}(c_j)\big)
\]
where $i$ indexes therapist turns and $j$ ranges over user turns in the look-ahead window $[i, i{+}k)$, with $k=2$. The window looks forward, following the nCLiD formulation~\cite{nasir2019modeling}, because a therapist's turn responds to what came before, so coordination may only surface later in the turn.

Transcript-level nCLiD averages these distances and normalizes by $\alpha$,
the mean cosine distance across all turn pairs in the transcript, which
corrects for within-speaker consistency and topic-driven similarity that
would otherwise inflate entrainment estimates:
\begin{equation}
    \mathbf{nCLiD} = \frac{\frac{1}{N}\sum_{i=1}^{N} d_i^{C \to A}}{\alpha}
\end{equation}
with $N=10$ turn pairs per transcript. Lower nCLiD indicates greater linguistic alignment with the user. We treat nCLiD as complementary to our valence--arousal trajectories, since semantic similarity captures linguistic convergence but not emotional tone or its shifts.

\subsubsection{\textbf{Protocol Leverage Force}}
\label{sec:force}

As model deployments scale, evaluation increasingly requires surfacing behavior-level patterns that aggregate metrics cannot capture~\cite{jones2025forecasting}. Following recent work showing that accuracy and F1 metrics can reward confident guessing over admitting uncertainty~\cite{kalai2026evaluating}, we favor behavior-level metrics. Thus, instead of  scoring correctness, we summarize the per-turn effect of the CBT-guided framework with a scalar we call \emph{Protocol Leverage Force}, denoted $\mathcal{F}$, a measure of behavioral reorientation, not therapeutic quality. The name reflects what the quantity measures: the mechanical advantage of a structured intervention, i.e.\ how much response reorientation is achieved per unit of deviation from the model's native response manifold. Inspired by the ``physics of language models'' perspective that studies LLM behavior as a dynamical system~\cite{AllenZhu-icml2024-tutorial}, and following Huygens's classical centrifugal force $F = mv^2/r$~\cite{huygens1980horologium}, we formally define $\mathcal{F}$ as follows:
\begin{equation}
\mathcal{F} = \frac{w_{\text{protocol}} \cdot \Delta^2}{d_{\cos}(e_{\text{MCoT}},\, \mu_{\text{base}}) + \varepsilon},
\label{eq:leverage}
\end{equation}
\begin{equation*}
\Delta = d_{\cos}(\text{baseline},\, \text{patient}) - d_{\cos}(\text{MCoT},\, \text{patient}),
\label{eq:delta}
\end{equation*}
where each factor captures a distinct dimension of the intervention:~$\mathbf{w_{\text{protocol}}}$: the weight of the CBT principle \textit{(V, SQ, AP)} selected on this turn. A confidently chosen protocol contributes more than an ambivalent one; this term reflects the framework's decision. $\mathbf{\Delta}$: the signed reorientation gain. Positive $\Delta$ means the MCoT response moved closer to the patient's surfaced cognitive state than the baseline would have. Squaring it ensures that the magnitude of reorientation dominates the signal while the sign is preserved separately for interpretation. $\mathbf{d_{\cos}(e_{\text{MCoT}}, \mu_{\text{base}}) + \varepsilon}$: How far the MCoT response has strayed from the centroid of baseline responses in embedding space. $\varepsilon$ is the \emph{baseline perturbation}: the mean distance between baseline centroids across three paraphrases of the baseline prompt, to test sensitivity to prompt wording.

$\mathcal{F}$ is large when a committed protocol produces a decisive reorientation toward the patient without the response drifting far from what the model would natively say. $\mathcal{F}$ is small under three distinct failure modes, each diagnosable from its terms: (i) low $w_{\text{protocol}}$: the framework was uncertain which strategy to apply; (ii) low $\Delta$: the response barely moved the dialogue relative to baseline; (iii) high denominator with moderate numerator: the model reframed stylistically but did not convert that distance into therapeutic gain. Reading the three terms jointly therefore exposes \emph{why} a turn succeeded or failed, not only whether it did. We additionally examine $\Delta$ in isolation, and by applying Otsu's method~\cite{otsu1979threshold} per model and CBT principle to avoid an arbitrary cutoff, separating high- and low-shift turns. These are reported in Appendix~\ref{app:otsu} and Table~\ref{tab:otsu_shift}.

\begin{figure}[t]
    \centering
    \small
    \includegraphics[width=0.8\columnwidth]{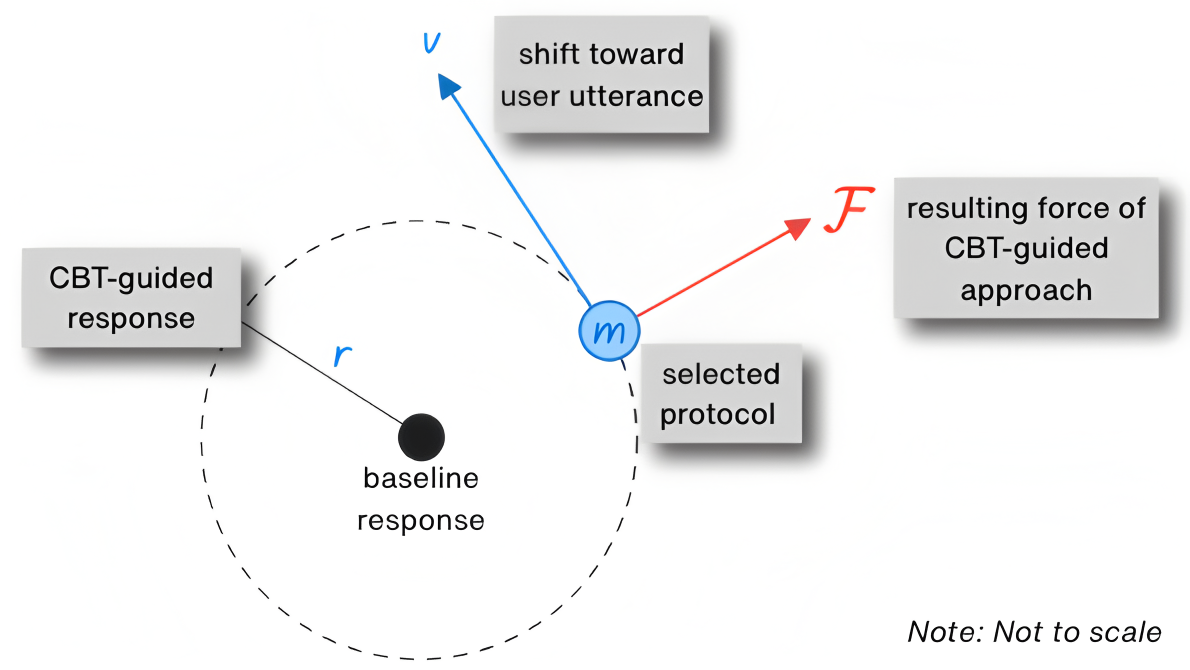}
    \caption{\small
    Centrifugal-force-inspired view of $\mathcal{F}$, following Huygens's $F = m v^{2}/r$~\cite{huygens1980horologium}. The selected protocol is a point mass $m$ orbiting the baseline centroid (the ``gravity'' of the model's default behavior) at radius $r = d_{\cos}(\mathbf{e}_{\text{MCoT}}, \boldsymbol{\mu}_{\text{base}}) + \varepsilon$. The tangential arrow $v$ is the signed shift of the response toward the user's utterance; the outward arrow is the resulting force $\mathcal{F}$, which grows when the CBT-guided approach moves the response away from baseline and toward the user.}
    \label{fig:force}
\end{figure}

\begin{table}[t]
\centering
\footnotesize
\setlength{\tabcolsep}{3pt}
\renewcommand{\arraystretch}{1.2}
\caption{Comparing baseline and MCoT examples for $\mathcal{F}$ (Eq.~\eqref{eq:leverage}). Both models receive the \emph{same} user query (``I'm extremely good at my job. I mean, I'm the best employee that they have ever had... I make no errors. So how can I be wrong when it comes to these circumstances with my coworkers?''). Red marks early assumptions dropped in MCoT; green marks the new shift MCoT introduces. Orange marks nearly unchanged phrasing.}
\label{tab:f_example}
\begin{tabular}{@{}p{0.16\columnwidth} p{0.60\columnwidth} r@{}}
\toprule
\textbf{Model/Mode} & \textbf{Response} & $\boldsymbol{\mathcal{F}}$ \\
\midrule
Gemma3\newline\textit{Baseline} & It's fantastic you're so confident and skilled in your
role, that's a wonderful asset! However, even the most talented individuals
can have blind spots when it comes to interpersonal relationships. \textcolor{red!80}{It's possible
your perception of the situation with your coworkers is influenced by your own
perspective and values}, and exploring that might reveal areas for understanding
or adjustment. & \\
\addlinespace[2pt]
Gemma3\newline\textit{MCoT (SQ)} & It sounds like you have a very strong sense of your
abilities and contributions at work. \textcolor{green!40!black}{Can you think of a time when you've
received feedback that suggested there might be room for improvement, even if it
was minor?} & $1.647\%$ \\
\midrule
Mistral\newline\textit{Baseline} & While it's great to hear about your dedication and high
performance, it's important to remember that everyone makes mistakes sometimes,
and it's a part of the learning process. \textcolor{orange!70!black}{Regarding your co-workers, maintaining
positive relationships is crucial for a harmonious work environment. Let's
discuss potential strategies for handling challenging situations with them.} & \\
\addlinespace[2pt]
Mistral\newline\textit{MCoT (AP)} & I understand that you feel very confident in your abilities
and believe that you're the best employee. \textcolor{orange!70!black}{However, it's important to consider
other perspectives as well. Let's explore how a trusted friend or colleague
might view this situation.} & $0.002\%$ \\
\bottomrule
\end{tabular}
\end{table}

\section{\textbf{Results \& Discussions}}
\label{sec:results}

\subsection{Linguistic Entrainment \& Valence--Arousal Trajectory}
\begin{figure*}[h]
\centering
\small
\includegraphics[width=\textwidth]{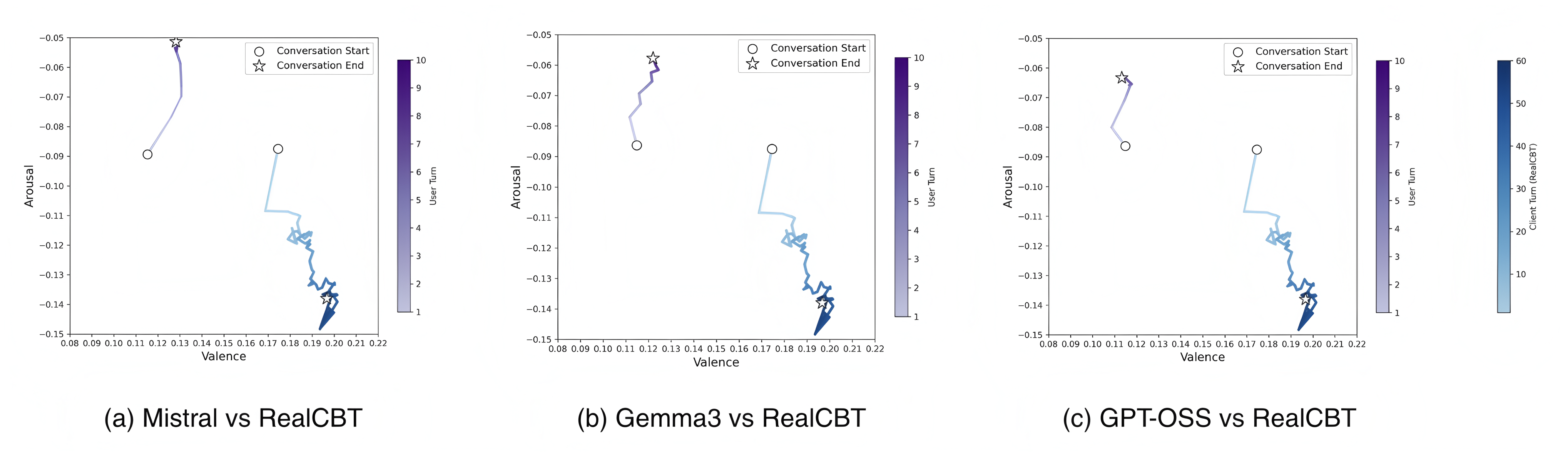}
\caption{\small Valence–Arousal trajectories of user language across CBT-MCoT synthetic sessions (purple) and RealCBT human transcripts (blue), scored using the NRC VAD Lexicon. Each point is a cumulative mean over turns, and each graph shows the average arc per model across 14 transcripts. RealCBT's longer sessions reveal a more nuanced emotional journey (more oscillations).}
\label{fig:va_all}
\end{figure*}

Across both synthetic and RealCBT sessions (Fig.~\ref{fig:va_all}), valence rises over time, marking a shift toward positive affect. Arousal tells a different story. RealCBT's 60-turn sessions de-escalate steadily, while our 10-turn synthetic sessions drift upward instead of calming. RealCBT's longer arcs are also richer, oscillating more than the synthetic ones. We read this divergence, not as a failure of the synthetic sessions to be lively, but as evidence that they capture the direction of valence change without the sustained de-escalation and affective depth of real therapy that Wang et al.\ document~\cite{wang2025realcbt}.

Table~\ref{tab:nclid} reports nCLiD across models. \texttt{GPT-OSS} scores lowest, followed by \texttt{Gemma3} and \texttt{Mistral}, and \textit{(V)} yields the highest entrainment in every model. Strikingly, our models entrain more than RealCBT does, the reverse of what Kian et al.\ found for \texttt{GPT-3.5-turbo}, which entrained less than human therapists~\cite{kian2025using}. The tension this reveals is telling. Synthetic sessions lean on \textit{(V)}, which mirrors the user's language into smoother, tighter arcs, while RealCBT's oscillations likely stem from therapists' heavier use of \textit{(SQ)}. In short, \textit{(V)} maximizes surface entrainment, but \textit{(SQ)} is what produces the affective arcs characteristic of real CBT.

\begin{table}[h]
\centering
\small
\setlength{\tabcolsep}{4pt}
\renewcommand{\arraystretch}{1.0}
\caption{
\small
\textsc{Linguistic entrainment per model.} \textit{Overall} is the dialogue-level 
nCLiD (mean across transcripts); \textit{V, SQ, AP} columns report nCLiD per principle. RealCBT (overall nCLiD 0.2752) is the CBT--dialogue reference. Lower nCLiD indicates greater entrainment highlighted in green.} 
\label{tab:nclid}
\begin{tabular}{lcccc}
\toprule
\textbf{Model} & \textbf{Overall} & \textbf{V} & \textbf{SQ} & \textbf{AP} \\
\midrule
Mistral & 0.2264 & 0.2061 & 0.2478 & \cellcolor{green!20}0.2327 \\
Gemma3   & 0.2015 & 0.1744 & 0.2062 & 0.2608 \\
GPT-OSS     & \cellcolor{green!20}0.1788 & \cellcolor{green!20}0.1743 & \cellcolor{green!20}0.1947 & -- \\
\bottomrule
\end{tabular}
\end{table}

\subsection{Linguistic Shift \& Therapeutic Force}

\subsubsection{\textbf{Validation and Socratic Questioning}}

Otsu-thresholded shift magnitudes and flagged counts (Appendix~\ref{app:otsu},
Table~\ref{tab:otsu_shift}) yield two rankings that, read together, separate
CBT knowledge (\texttt{GPT-OSS~$>$~Gemma3~$>$~Mistral}) from CBT applicability
(\texttt{Mistral~$=$~Gemma3~$>$~GPT-OSS}): \texttt{GPT-OSS} carries the largest
per-model shift distributions, while \texttt{Mistral} and \texttt{Gemma3} redirect
most consistently, chiefly under \textit{(V)}. The shift is noticeable across all
three models. Table~\ref{tab:wilcoxon_shift} confirms the pattern statistically:
each model's default-routed protocol \texttt{(Mistral-V, Gemma3-SQ, GPT-OSS-SQ)}
shows no significant shift ($p > 0.05$), while non-default protocols shift in ways
consistent with their therapeutic intent: \textit{(V)} pulls the response closer to
the patient (negative shift), while \textit{(SQ)} pushes it further (positive shift).

\begin{table}[h]
\centering
\setlength{\tabcolsep}{4pt}
\renewcommand{\arraystretch}{1.0}
\caption{
\small
\textsc{Wilcoxon signed-rank test on signed shifts.} Per-turn signed shifts $\Delta$, testing $H_0\colon\text{median shift}=0$ per (model, protocol) cell. Shift $=d_{\cos}(\text{MCoT},\text{patient})-d_{\cos}(\text{baseline},\text{patient})$. Highlighted cells indicate significant shifts at $p<.001$.}
\label{tab:wilcoxon_shift}
\small
\setlength{\tabcolsep}{3.5pt}
\begin{tabular}{llrrrrr}
\toprule
\textbf{Model} & \textbf{Proto.} & $\boldsymbol{n}$ & \textbf{Median} & \textbf{Mean} & $\boldsymbol{W}$ & $\boldsymbol{p}$ \\
\midrule
\multirow{3}{*}{Mistral}
  & V   &  90 & $+0.0030$ & $-0.0001$ & 1991.00 & $.820$ \\
  & SQ  &  36 & $+0.1040$ & $+0.1051$ &   86.00 & \cellcolor{green!20}$<.001$ \\
  & AP  &  14 & $+0.1187$ & $+0.0882$ &   19.00 & $.035$ \\
\midrule
\multirow{3}{*}{Gemma3}
  & V   &  30 & $-0.0962$ & $-0.1127$ &   14.00 & \cellcolor{green!20}$<.001$ \\
  & SQ  & 109 & $-0.0026$ & $+0.0064$ & 2906.00 & $.782$ \\
  & AP  &   1 & $+0.1839$ & $+0.1839$ &    0.00 & $1.000$ \\
\midrule
\multirow{2}{*}{GPT-OSS}
  & V   & 115 & $-0.0435$ & $-0.0534$ & 1771.00 & \cellcolor{green!20}$<.001$ \\
  & SQ  &  25 & $+0.1141$ & $+0.0680$ &   94.00 & $.067$ \\
\bottomrule
\end{tabular}
\end{table}

\subsubsection{\textbf{Alternative Perspective}}

\begin{table}[h]
\centering
\small
\setlength{\tabcolsep}{4pt}
\renewcommand{\arraystretch}{1.0}
\caption{
\small
\textsc{Transcript-level impact of MCoT via $\mathcal{F}$.} $\varepsilon$: mean baseline perturbation across three prompt paraphrases (small values indicate responses are not sensitive to paraphrasing).}
\begin{tabular}{lccc}
\toprule
\textbf{Model} &  \textbf{Mean} & \textbf{Standard deviation} & $\boldsymbol{\varepsilon}$ \\
\midrule
Mistral  & 1.256\% & 0.442\% & 0.040 \\
Gemma3  & 1.179\% & 0.690\% & 0.061 \\
GPT-OSS  & 1.341\% & 0.500\% & 0.053 \\
\bottomrule
\end{tabular}
\label{tab:force}
\end{table}

\textit{(AP)} behaves differently. \texttt{GPT-OSS} and \texttt{Gemma3} route to
\textit{(AP)} too rarely to threshold, leaving \texttt{Mistral} as the only model
flagged under it (Appendix~\ref{app:otsu}). The empty cells indicate models forgo \textit{(AP)} in favor of their preferred protocols.

\subsubsection{\textbf{Protocol Leverage Force}}
Table~\ref{tab:force} reports the transcript-level $\mathcal{F}$ score per model. Each model's average $\mathcal{F}$ falls within a narrow $1.18\%$--$1.34\%$ band, a clear signal that knowledge-guided prompting nudges these models but cannot overcome their underlying bias. \texttt{Gemma3} shows the most dialogue-to-dialogue variability, and \texttt{Mistral} the least. Nor are these low scores merely noise, since Table~\ref{tab:wilcoxon_shift} confirms the framework moves responses in the intended direction, pulling \textit{V} toward the user's language and pushing \textit{SQ} away from it, even as $\mathcal{F}$ reveals how modest that movement is. The two tables show that protocol adherence is real but insufficient to overcome the models' proclivity toward \textit{V}. Response quality is inherently subjective, as our low inter-rater agreement confirms. We therefore treat $\mathcal{F}$ as a complement to expert evaluation, not a replacement for it. This points to a future direction towards mechanistic validation that could trace how guidance reshapes a model's internal pathways \cite{aggarwal2026moral}.

\subsection{Human Evaluations}

\subsubsection{\textbf{Inter-Rater Agreement on CBT Principles}}
Table~\ref{tab:partA_distributions} reports human experts' confidence 
and inter-rater agreement per CBT principle. Krippendorff's $\alpha$ measures rater consistency across the full confidence scale, where 
$\alpha = 0$ indicates chance and $\alpha = 1$ perfect agreement 
\cite{krippendorff2004content}. Two key observations emerge. 
First, \textit{(V)} consistently achieved the highest confidence yet the lowest $\alpha$, indicating that while raters recognized \textit{(V)}, they disagreed on its quality, reflecting the model's inherent tendency toward empathetic acknowledgment and suggesting \textit{(V)} is underspecified as an evaluation target. Second, \textit{(SQ)} showed lower confidence but the highest inter-rater agreement (\texttt{Mistral}, $\alpha = 0.43$), indicating that although applied more selectively, it provides a more reliable and consistently interpreted evaluation signal among human experts.

\subsubsection{\textbf{Inter-Rater Agreement on Protocol Preference}}
As shown in Table~\ref{tab:protocol_summary}, expert evaluation is highly subjective: Fleiss' Kappa ($\kappa$) stays near 0 across all models (near-chance agreement). \textit{Keep} refers to cases where the MCoT-selected response is preferred, \textit{Other} indicates preference for responses generated by the other principles, and \textit{Neither} denotes that neither option is preferred. \texttt{Gemma3} achieves the highest $\kappa$ (0.123) and \textit{Keep} rate (57.5\%), indicating that its MCoT-selected responses are most often preferred, consistent with its stronger performance under (LLM-as-a-Judge) LaaJ in the detailed Study provided in (Appendix~\ref{tab:llm_judge_results}). In contrast, \texttt{Mistral} shows the lowest \textit{Keep} rate (22.2\%) and highest \textit{Other} rate (58.7\%), suggesting weaker alignment between its selections and expert preference. This low agreement supports our central claim. Raters were blind to the applied intervention yet still could not agree on the best strategy for a query, indicating that judging therapeutic appropriateness is inherently subjective when models produce empathetically fluent but strategically ambiguous responses.

\begin{table}[h]
\small
\caption{
\small
\textsc{Reviewer confidence per CBT principle.} Percent of ratings at 3 or 4 on the 0--4 scale. Krippendorff's $\alpha$ (ordinal) measures inter-rater agreement on the full scale. ``Not Applicable'' responses were recoded as 0.}
\label{tab:partA_distributions}
\centering
\setlength{\tabcolsep}{4pt}
\renewcommand{\arraystretch}{1.0}
\begin{tabular}{l c c c}
\toprule
\textbf{Model} & \textbf{Principle} & \textbf{Confidence (\%)} & $\boldsymbol{\alpha}$ \\
\midrule
\multirow{3}{*}{Mistral}
  & V  & \cellcolor{green!20}77.8 & 0.20 \\
  & SQ & 11.1 & \cellcolor{green!20}0.43 \\
  & AP & 27.0 & 0.41 \\
\midrule

\multirow{3}{*}{Gemma3}
  & V  & \cellcolor{green!20}63.8 & \cellcolor{red!20}-0.01 \\
  & SQ & 52.5 & \cellcolor{green!20}0.35 \\
  & AP &  7.5 & 0.22 \\
\midrule
\multirow{3}{*}{GPT-OSS}
  & V  & 46.3 & 0.06 \\
  & SQ & \cellcolor{green!20}61.3 & \cellcolor{green!20}0.10 \\
  & AP & 35.0 & 0.01 \\
\bottomrule
\end{tabular}
\end{table}

\begin{table}[h]
\caption{
\small
\textsc{Expert principle preference.}
Whether experts preferred to \textit{Keep} the MCoT-selected principle, select an alternative (\textit{Other}), or \textit{Neither}. Percentage of ratings per category across 3 transcripts, 10 turns, and 8 raters.}
\centering
\small
\setlength{\tabcolsep}{4pt}
\renewcommand{\arraystretch}{1.0}
\begin{tabular}{lcccc}
\toprule
\textbf{Model} & \textbf{Keep (\%)} & \textbf{Other (\%)} 
& \textbf{Neither (\%)} & \textbf{$\kappa$} \\
\midrule
Mistral & 22.2 & \cellcolor{red!20}58.7 & 19.0 & 0.099 \\
Gemma3   & \cellcolor{green!20}57.5 & 32.5 & 10.0 & \cellcolor{green!20}0.123 \\
GPT-OSS     & \cellcolor{green!20}47.5 & 26.2 & 26.2 & 0.024 \\
\bottomrule
\end{tabular}
\label{tab:protocol_summary}
\end{table}

\section{\textbf{Conclusion \& Limitations}}
We present a knowledge-guided CBT reasoning framework that grounds LLM responses in structured clinical knowledge, evaluated through expert judgment, linguistic entrainment, and our proposed Protocol Leverage Force ($\mathcal{F}$), a novel metric for gauging CBT-guided intervention beyond the surface polish of a response. Our approach yields consistent but modest gains over baseline. \texttt{Gemma3} leads the pack, with its MCoT-selected responses being the ones reviewers most often chose to keep. Both $\mathcal{F}$ and entrainment tell the same story: prompt guidance pulls responses in the right direction, but not enough to move the models away from \textit{(V)}, and reach more towards \textit{(SQ)} or \textit{(AP)}, an empathy bias that prior work has already documented \cite{tak2025aware}. Two key limitations temper these findings. First, the PBCCD surfaces unhelpful cognitions but not the client's strengths or adaptive resources; integrating the Strength-Based Cognitive Conceptualization Diagram (SBCCD) to model both remains future work. Second, our evaluation, too, is bounded, drawn from 14 synthetic, RealCBT-derived case studies with LLM-simulated user turns, deliberately excluding clear of self-harm and other safety-critical cases; widening this to a fuller range of clinical issues is likewise left for later. Our deeper contribution to affective computing is not a technique but an instrument, a means of asking whether structured guidance changes what a model does, not merely what it knows. Knowing CBT is not the same as practicing it. Closing the gap that $\mathcal{F}$ makes measurable is the next step toward affect-aware systems that reason about feeling, rather than merely respond.

\clearpage
{\section*{\textbf{Ethical Impact Statement}}
This work is a research prototype and is not currently intended for  deployment. All user dialogues used in evaluation are synthetically generated; no real patient data, therapy transcripts, or personally identifiable information were used at any stage. Evaluation is conducted for 14 user case studies, with 
user turns in both studies simulated by \texttt{GPT-4o-mini}. All local model inference (\texttt{GPT-OSS-20B}, \texttt{Mistral-7B}, \texttt{Gemma3-12B}) was conducted on a single NVIDIA  GeForce RTX 5090 GPU, with the full evaluation pipeline completing in approximately 5 hours; we report this to support reproducibility and to document the compute footprint of our study.

The RealCBT dataset~\cite{wang2025realcbt}, from which user case studies were constructed, comprises authentic CBT dialogues transcribed from publicly available video-sharing platforms, including YouTube and Vimeo. These recordings consist of role-play demonstrations and educational sessions featuring therapist-client simulations, rather than recordings of 
real clinical sessions with actual patients. No additional consent or data collection  was conducted by the present study beyond the use of published, openly available material. While the source recordings contain no direct personal identifiers and the user turns in our evaluation are synthetically generated, we acknowledge that publicly posted therapeutic role-play content can still carry residual reidentification risk through voice, visual, or contextual cues present in the original videos; our pipeline operates only on transcribed text and does not propagate these signals, but downstream users of RealCBT-derived resources should apply additional safeguards such as restricted access and secure storage.

Study~2 comprises 14 user case-study queries expanded into 280 multi-turn dialogues (baseline and MCoT mode), spanning career, academic, relational, self-esteem, health, and financial distress, while excluding self-harm risk and harm toward others. This breadth captures meaningful variation in the concerns users bring to such a system, but remains a narrow evaluation window that does not span the full range of emotional trajectories, conversational dynamics, or escalation patterns seen in real therapeutic practice. Findings should therefore be read as indicative of relative LLM behavior under structured conditions, not as evidence of clinical efficacy.

The use of LLM-as-a-Judge evaluation, employed in our preliminary study (Appendix~\ref{appendix:results}) alongside human expert assessment, should be interpreted as a scalable proxy for relative model comparison rather than as a substitute for clinical validation. For both studies, human evaluation surveys were administered anonymously; evaluators were not required to provide identifying information, and all responses were recorded and stored in de-identified form through the affiliated university's Qualtrics survey platform. Evaluators were compensated for their participation in recognition of the time and domain expertise required to assess multi-turn therapeutic dialogues. No content involving suicidal ideation was included in any human evaluation survey, in recognition of the psychological burden such material places on evaluators. Additionally, evaluators were informed about their right to opt-out at any time and that they would still receive their compensation.

Our framework contributes to efforts of decomposing the latent cognitive structure underlying user-expressed distress into clinically interpretable representations that can inform structured therapeutic intervention. While this has clear benefits for affective computing systems that rely on understanding a user's internal appraisal processes, such as mental health support tools and CBT-aligned dialogue agents, the capacity to infer automatic thoughts, emotional responses, and behavioral patterns from natural language raises important privacy considerations and warrants further scrutiny and empirical evaluation as an open challenge for future work in affective computing research. 

Although this work is a research prototype, we recognise that the gap between proof-of-concept affective computing systems and deployed mental health tools has narrowed considerably, and that deferring ethical reflection to future work is no longer tenable. If such a tool were deployed in a real-world setting, it is essential that users be informed about and consent to the functionality, the nature of cognitive inference being performed, and the potential risks of systems that model their internal psychological states. 
Such systems must not be deployed without the explicit knowledge and consent of those whose cognitive and affective processes are being interpreted. Unconsented application of cognitive decomposition models carries the risk that deeply personal information about a user's belief structures, emotional vulnerabilities, and coping behaviors may be involuntarily exposed or used in ways that compromise their psychological privacy and well-being, use cases we explicitly do not support. We therefore envision this framework not as a standalone tool but as a clinical aid deployed under the guidance and oversight of qualified practitioners, enabling therapists and mental health professionals to develop a richer, structured understanding of their clients' cognitive patterns.

Moreover, this framework does not currently include safeguards for crisis escalation and must not be made accessible to vulnerable populations without robust safety mechanisms and clear referral pathways to qualified professionals. Broader ethical questions of accountability, consent, and equitable access to mental
health AI remain open and require interdisciplinary engagement beyond the scope of this work.

\section*{\textbf{Acknowledgment}}
We thank ACII reviewers for their constructive feedback, and the Departments of Psychology and Information Systems at UMBC for their collaboration. This work was supported in part by a gift award from NeuralNest LLC and USISTEF: US-India Technology Endowment Fund. All the opinions are those of the authors and not of the sponsoring organizations.

\appendices
\clearpage

\renewcommand{\thetable}{\thesection\arabic{table}}
\renewcommand{\thefigure}{\thesection\arabic{figure}}
\counterwithin{table}{section}
\counterwithin{figure}{section}

\section{System Prompts}
\label{appendix:prompts}

\begin{promptbox}{User Cognitive Model}
You are a clinical-reasoning assistant trained to extract a structured CBT cognitive model
from user text without diagnosing.

\medskip
\textbf{Field definitions:}
\begin{itemize}
  \item \textbf{Triggers:} External situations or contexts that precede distress (not thoughts or beliefs).
  \item \textbf{Automatic thoughts:} Immediate internal thought interpretations or predictions.
  \item \textbf{Emotions:} Momentary affective states.
  \item \textbf{Behaviors:} Observable actions or avoidance responses.
\end{itemize}

\medskip
Returns a JSON object with fields: \texttt{\{triggers, automatic\_thoughts, emotions, behaviors\}}.
\end{promptbox}

\medskip

\begin{promptbox}{CBT-Guided Multiple-Chains-of-Thought}
You will receive:
\begin{enumerate}
  \item \textbf{User Cognitive Model:} a structured decomposition of the user's experience.
  \item \textbf{Clinical Context:} clinical statements related to the user's language.
  \item \textbf{CBT Protocol Guideline:} describes three intervention strategies.
\end{enumerate}

Before generating the final response, internally simulate three candidate
responses (one per CBT intervention protocol). For each candidate:
\begin{enumerate}
  \item Infer the belief or assumption driving the user's experience.
  \item Apply the intervention principle using its techniques.
  \item Consider how the user might respond to that intervention.
\end{enumerate}

\textbf{Intervention candidates:}
\begin{itemize}
  \item \textbf{Validate \& Reflect:} Focus on emotional acknowledgment and alignment.
  \item \textbf{Socratic Questioning:} Ask questions that help the user examine the belief.
  \item \textbf{Alternative Perspective:} Introduce a gentle alternative interpretation of the belief.
\end{itemize}

\medskip
Evaluate and score which candidate would most effectively move the conversation forward
therapeutically. \textbf{Select the strongest candidate.}
\end{promptbox}

\section{Additional Results}
\label{appendix:results}

This appendix provides the complete empirical results from our preliminary study, which established the foundation for the knowledge-guided CBT reasoning framework. These tables detail the performance across four open-weight models using both automated and human-expert evaluation metrics.

\subsection{\textbf{Human Expert Evaluation.}}
Table~\ref{tab:human_eval} summarizes human expert assessments of protocol adherence and effectiveness.
\begin{table}[h]
\centering
\caption{\textsc{Human expert evaluation results.} Adherence captures the distribution of CBT strategies (V, SQ, CR, O/N), while effectiveness reflects appropriateness of the response for the applied principle (Likert Scale 1–5).}
\label{tab:human_eval}

\scriptsize
\setlength{\tabcolsep}{4.5pt}
\renewcommand{\arraystretch}{1.15}

\resizebox{\columnwidth}{!}{
\begin{tabular}{llcccccc}
\toprule
\textbf{Model} & \textbf{Mode} 
& \multicolumn{4}{c}{\textbf{Protocol Adherence (\%)}} 
& \multicolumn{2}{c}{\textbf{Protocol Effectiveness}} \\
\cmidrule(lr){3-6} \cmidrule(lr){7-8}

& & \textbf{V} & \textbf{SQ} & \textbf{CR} & \textbf{O/N} 
& \textbf{Mean} & \textbf{Var} \\
\midrule

Mistral & MCoT 
& 56.3 & 25.4 & 15.5 & 2.8 
& 3.80 & 0.51 \\

DeepSeek-R1-8B & CoT 
& \cellcolor{green!20}66.7 & 7.4 & \cellcolor{green!20}24.1 & 1.9 
& 3.53 & 0.69 \\

Gemma3 & MCoT 
& 54.5 & 36.4 & 6.1 & 3.0 
& \cellcolor{green!20}4.11 & \cellcolor{green!20}0.32 \\

GPT-OSS & CoT 
& 37.9 & \cellcolor{green!20}37.9 & 18.4 & \cellcolor{green!20}5.8
& 3.94 & 0.66 \\

\bottomrule
\end{tabular}
}

\vspace{2pt}

{\footnotesize
$\alpha_{\text{effectiveness}} = 0.39$, $\alpha_{\text{adherence}} = 0.44$. 
\texttt{Gemma} vs \texttt{DeepSeek}: $p = 0.002$ (significant); others: $p > 0.05$. Green highlights indicate the highest adherence per strategy and the highest protocol effectiveness.
}
\end{table}

\subsection{\textbf{LLM-as-a-Judge (LaaJ).}}
In Table~\ref{tab:llm_judge_results}, across all prompting modes, LaaJ results (\texttt{GPT-5.1}) consistently rank models as: \texttt{Gemma3-12B} $>$ \texttt{GPT-OSS-20B} $>$ \texttt{DeepSeek-R1-8B} $>$ \texttt{Mistral-7B}.

\begin{table}[h]
\centering
\caption{\textsc{LLM-as-a-Judge evaluation for protocol effectiveness.} Results from GPT-5.1 and Qwen3-32B judges across model responses and prompting modes based on our defined CBT-guided intervention techniques rubric. Effectiveness denotes the percentage of appropriate responses; agreement reflects inter-judge consistency.}
\label{tab:llm_judge_results}
\setlength{\tabcolsep}{4pt}
\begin{tabular}{llccc}
\toprule
& & \multicolumn{3}{c}{\textbf{Protocol Effectiveness(\%)}} \\
\cmidrule(lr){3-5}
\textbf{Response Model} & \textbf{Mode}
& \textbf{GPT-5.1} & \textbf{Qwen3-32B} & \textbf{Agreement(\%)} \\
\midrule

\multirow{3}{*}{Mistral}
& Baseline & 37.8 & 43.6 & \cellcolor{green!20}94.2 \\
& CoT  & \cellcolor{green!20}48.8 & \cellcolor{green!20}55.8 & 93.0 \\
& MCoT & 44.4 & 52.8 & 91.6 \\

\midrule
 
\multirow{3}{*}{DeepSeek-R1-8B}
& Baseline & 50.0 & 56.8 & 93.2 \\
& CoT  & 54.8 & \cellcolor{green!20}62.0 & 92.8 \\
& MCoT & \cellcolor{green!20}55.2 & 57.4 & \cellcolor{green!20}97.8 \\

\midrule

\multirow{3}{*}{Gemma3-12B}
& Baseline & 59.4 & 60.2 & \cellcolor{green!20}99.2 \\
& CoT  & 62.0 & 63.4 & 98.6 \\
& MCoT & \cellcolor{green!20}66.4 & \cellcolor{green!20}67.8 & 98.6 \\

\midrule

\multirow{3}{*}{GPT-OSS}
& Baseline & 51.8 & 58.2 & 93.6 \\
& CoT  & \cellcolor{green!20}58.8 & \cellcolor{green!20}71.8 & \cellcolor{red!20}87.0 \\
& MCoT & 58.0 & 62.0 & \cellcolor{green!20}96.0 \\

\bottomrule
\end{tabular}
\vspace{2pt}
{\footnotesize
Green highlights indicate highest protocol effectiveness per model (per judge) and highest agreement; red highlights indicate lowest agreement.

}

\end{table}

\begin{figure*}[!htbp]
\centering

\textbf{\small LaaJ Model: GPT-5.1}\\[-2pt]
\makebox[\textwidth][c]{%
\subfloat[Mistral-7B]{\includegraphics[width=0.25\textwidth]{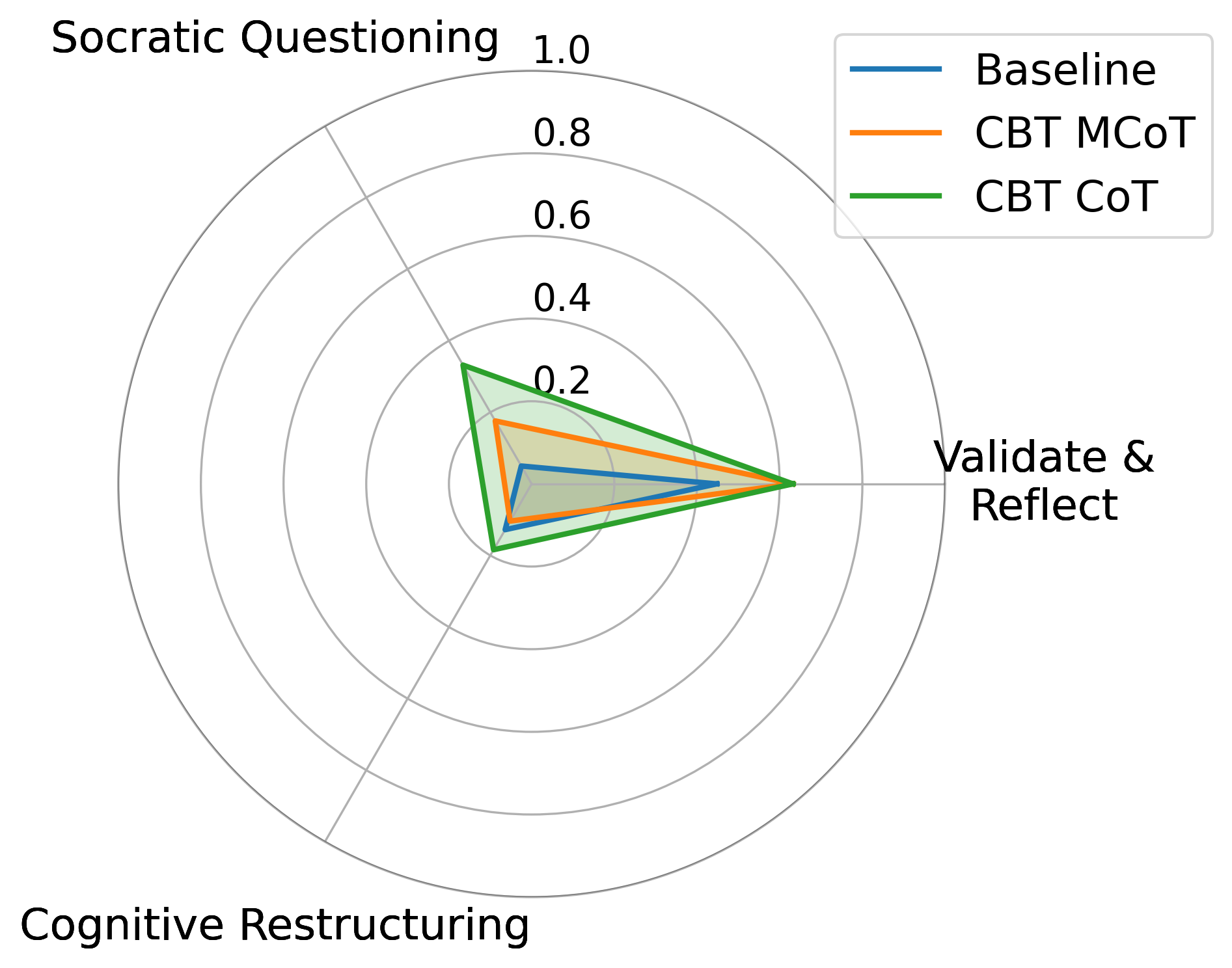}}
\subfloat[DeepSeek-R1-8B]{\includegraphics[width=0.25\textwidth]{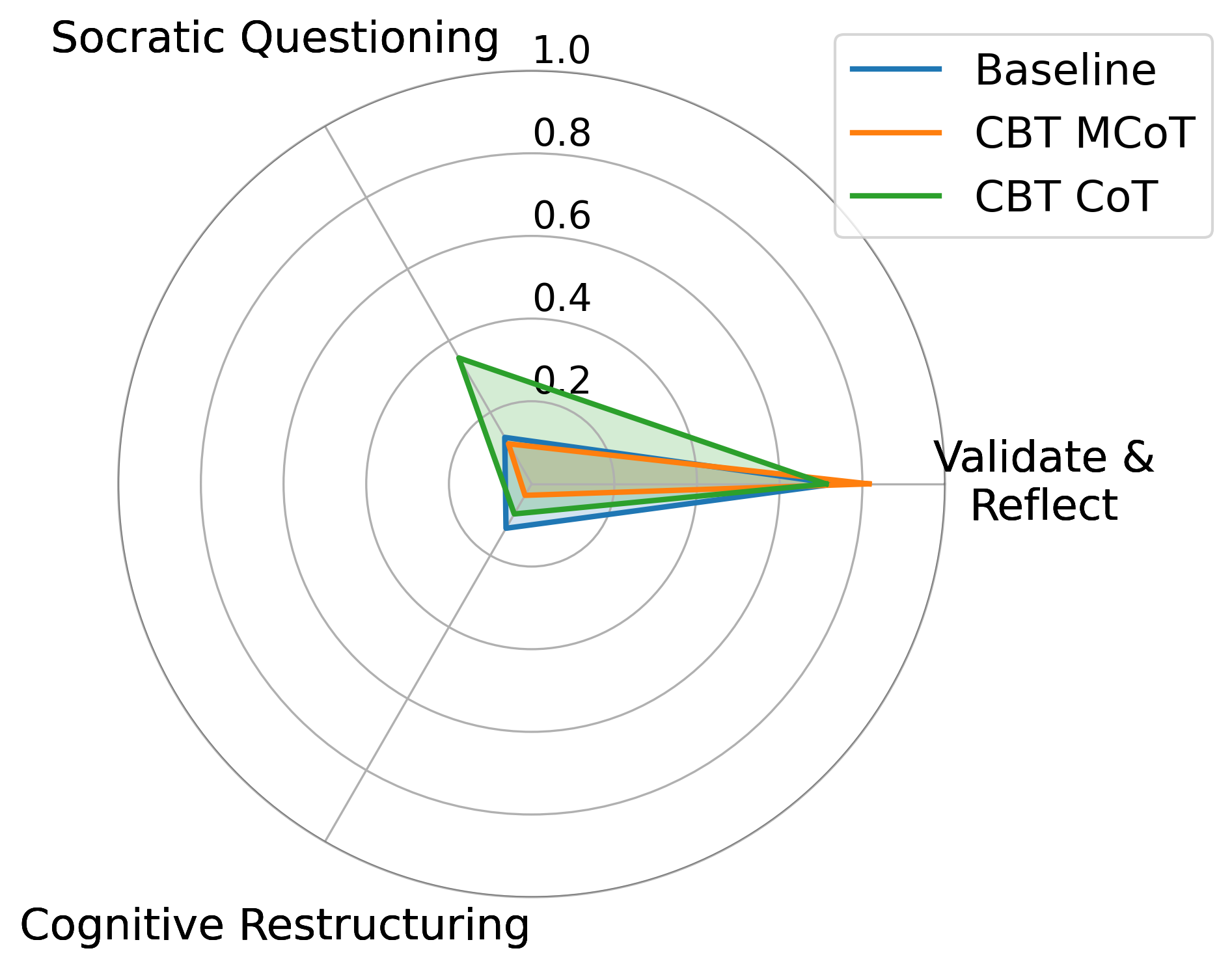}}
\subfloat[Gemma3-12B]{\includegraphics[width=0.25\textwidth]{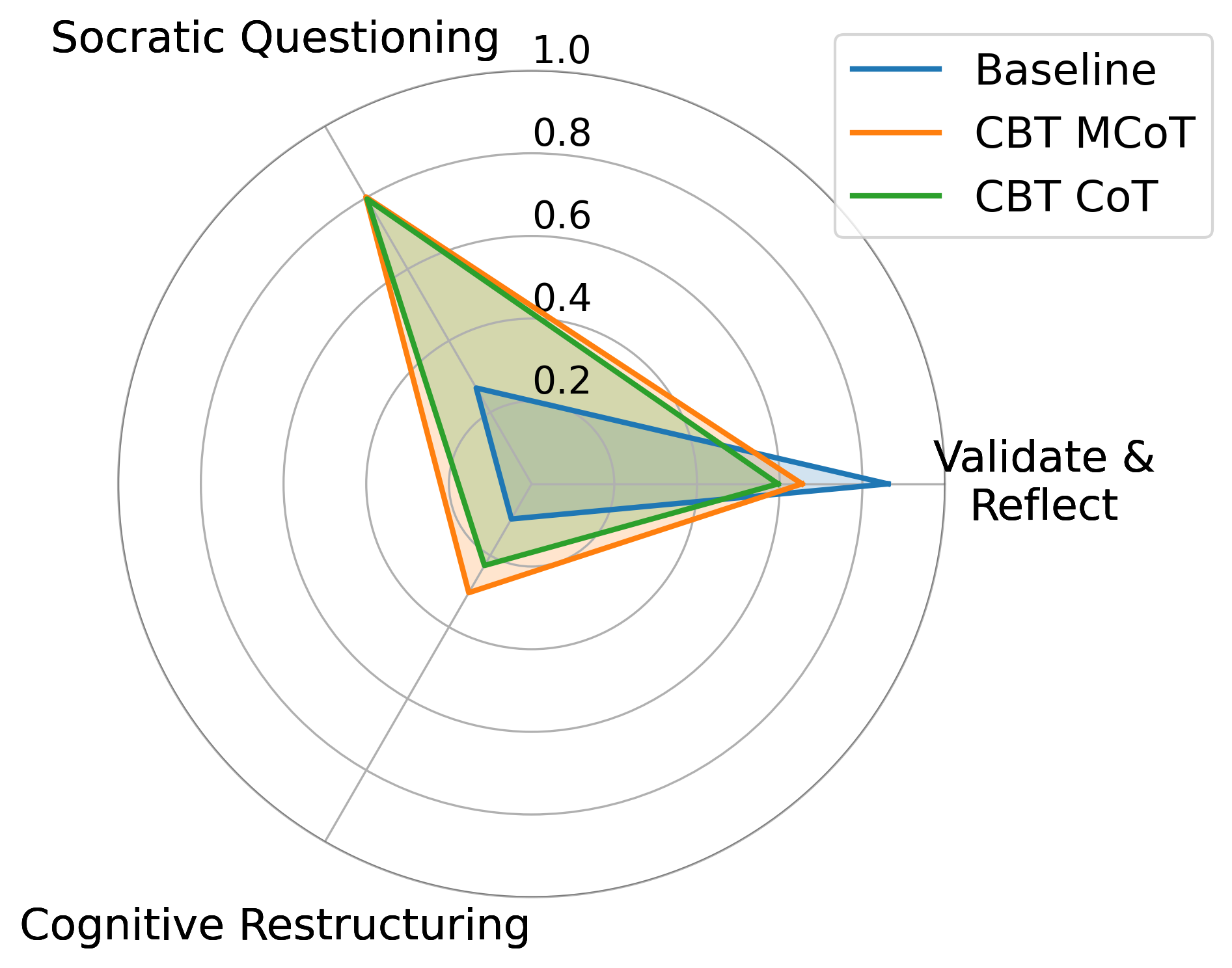}}
\subfloat[GPT-OSS-20B]{\includegraphics[width=0.25\textwidth]{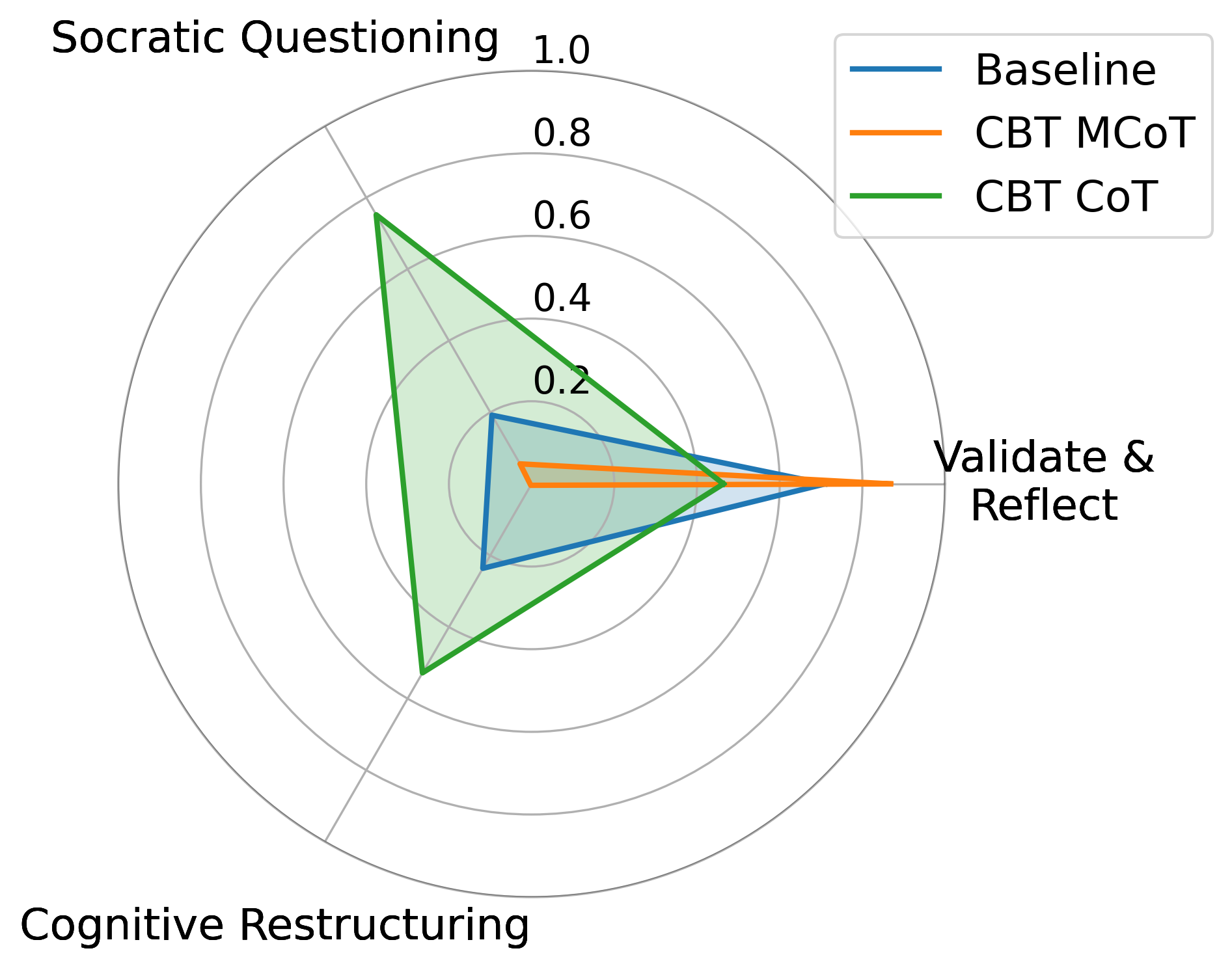}}
}\\[12pt]

\caption{CBT protocol adherence across response models under the GPT-5.1 judge. The radar plots illustrate the distribution of intervention strategies: Validation \& Reflection, Socratic Questioning, and Cognitive Restructuring across Baseline, CBT-CoT, and CBT-MCoT modes}

\label{fig:cbt_radar_judges}
\end{figure*}

\subsection{\textbf{Distribution of Concept Sources}}

In Table~\ref{tab:concept_sources}, we analyze concept usage during reasoning by attributing referenced concepts to either the \textit{user cognitive model}, \textit{retrieved SNOMED CT concepts}, or \textit{other/null}.

\begin{table}[!ht]
\centering
\caption{\textsc{Concept source distribution (\%).} Percentages indicate reliance on Clinical User Cognitive Model (Cog), SNOMED CT Concepts, or Other/Null sources.} 
\label{tab:concept_sources}
\footnotesize
\setlength{\tabcolsep}{6pt}
\renewcommand{\arraystretch}{1.1}

\begin{tabular}{lllc cccc}
\toprule
\textbf{Model} & \textbf{Mode} & \textbf{Intv.}
& \textbf{Cog} & \textbf{SNOMED} & \textbf{O} & \textbf{N} \\
\midrule

Mistral
& CoT  & All & \cellcolor{green!20}66.7 & 33.3 & 0.8 & -- \\
& MCoT & V   & 33.3 & 66.7 & -- & -- \\
&      & SQ  & 31.1 & \cellcolor{green!20}68.9 & -- & -- \\
&      & CR  & 51.8 & 48.2 & -- & -- \\

\midrule

DeepSeek-R1-8B 
& CoT  & All & \cellcolor{green!20}67.9 & 23.9 & \cellcolor{green!20}4.4 & 3.8 \\
& MCoT & V   & 33.3 & \cellcolor{green!20}53.7 & 1.6 & 11.4 \\
&      & SQ  & 62.3 & 26.2 & 2.3 & 9.2 \\
&      & CR  & 11.1 & 0.0 & -- & \cellcolor{red!20}88.9 \\

\midrule

Gemma3
& CoT  & All & \cellcolor{green!20}67.5 & 21.1 & 2.6 & \cellcolor{green!20}8.8 \\
& MCoT & V   & 61.5 & \cellcolor{green!20}36.0 & 0.5 & 2.0 \\
&      & SQ  & 56.2 & 30.4 & \cellcolor{green!20}8.2 & 3.6 \\
&      & CR  & 61.8 & 28.8 & 7.5 & 2.4 \\

\midrule

GPT-OSS
& CoT  & All & 30.5 & 47.7 & \cellcolor{green!20}12.5 & 7.8 \\
& MCoT & V   & 39.0 & 40.3 & 5.2 & 14.9 \\
&      & SQ  & \cellcolor{green!20}43.3 & \cellcolor{green!20}48.1 & 4.8 & 3.8 \\
&      & CR  & 20.6 & 43.1 & 2.9 & \cellcolor{green!20}32.4 \\

\bottomrule
\end{tabular}

\vspace{2pt}
\footnotesize{Intv: Intervention strategy, Cog: Cognitive model, O: Other, N: Null. Green highlights indicate dominant concept source per setting; red highlights indicate high reliance on Null (unsupported) usage.}
\end{table}

\subsection{\textbf{Alignment between retrieved evidence and concept usage in model reasoning.}}
Table~\ref{tab:alignment} compares the distribution of retrieved concepts with those used during model reasoning. The \textbf{retrieved} columns show the distribution of concepts (entailment, neutral, contradiction) after NLI Classification

\begin{table}[!ht]
\centering
\caption{\textsc{Alignment between retrieved evidence and concept usage.} Retrieved \% denotes the top-5 retrieval classified by NLI into Entailment (E), Neutral (N), Contradiction (C). Used \% shows the concepts used from SNOMED CT (or other/null) in the reasoning trace of the model before response generation.}
\label{tab:alignment}

\scriptsize
\setlength{\tabcolsep}{2.5pt}
\renewcommand{\arraystretch}{1.1}

\resizebox{\columnwidth}{!}{
\begin{tabular}{llccc cccc}
\toprule
\textbf{Model} & \textbf{Mode} 
& \multicolumn{3}{c}{\textbf{Retrieved (\%)}} 
& \multicolumn{4}{c}{\textbf{Used (\%)}} \\
\cmidrule(lr){3-5} \cmidrule(lr){6-9}
& & E & N & C
  & E & N & C & (O+N) \\
\midrule

Mistral
& CoT  & 46.0 & \cellcolor{green!20}49.2 & 4.8 & \cellcolor{green!20}83.3 & 16.7 & 0.0 & 0.0 \\
& MCoT & 35.2 & \cellcolor{green!20}52.4 & 12.4 & \cellcolor{green!20}74.3 & 24.3 & 1.4 & 0.0 \\

\midrule

DeepSeek-R1-8B
& CoT  & 37.2 & \cellcolor{green!20}51.6 & 11.2 & \cellcolor{green!20}73.7 & 23.7 & 2.6 & 25.5 \\
& MCoT & 38.8 & \cellcolor{green!20}52.4 & 8.8  & \cellcolor{green!20}78.8 & 21.2 & 0.0 & 42.1 \\

\midrule

Gemma3 
& CoT  & 40.0 & \cellcolor{green!20}55.6 & 4.4 & \cellcolor{green!20}70.8 & 29.2 & 0.0 & 0.0 \\
& MCoT & 27.6 & \cellcolor{green!20}66.8 & 5.6 & \cellcolor{green!20}62.5 & 37.5 & 0.0 & 0.0 \\

\midrule

GPT-OSS 
& CoT  & 27.2 & \cellcolor{green!20}57.6 & 15.2 & \cellcolor{green!20}58.1 & 40.3 & 1.6 & 29.5 \\
& MCoT & 41.2 & \cellcolor{green!20}48.0 & 10.8 & \cellcolor{green!20}71.0 & 29.0 & 0.0 & 36.7 \\

\bottomrule
\end{tabular}
}
\vspace{2pt}
\footnotesize{E: Entail, N: Neutral, C: Contradict, O+N: Other + Null. Green highlights indicate highest values within each distribution (retrieved/used); Although more N (neutral) concepts were retrieved, models show tendency towards selecting a higher \% of E (entailment) concepts in the reasoning trace, implying stronger grounding alignment to concepts provided.}
\end{table}

\subsection{\textbf{Impact of SNOMED CT and Cognitive Model on Response Appropriateness}}

Table~\ref{tab:human_concept} evaluates how grounding sources influence model responses. Appropriateness is measured as the proportion of expert ratings labeled \emph{Appropriate} or \emph{Very Appropriate} across all dialogue turns.

\begin{table}[!ht]
\centering
\caption{\textsc{Human evaluation and concept usage.}
Appropriateness is the percentage of ratings labeled as \textit{Appropriate} or \textit{Very appropriate}. SNOMED CT and Clinical User Cognitive Model percentages reflect the distribution of concept types used in reasoning.}
\label{tab:human_concept}

\footnotesize
\setlength{\tabcolsep}{3pt}
\renewcommand{\arraystretch}{1.1}

\begin{tabular}{llccc}
\toprule
\textbf{Model} & \textbf{Mode} & \textbf{Effectiveness (\%)} & \textbf{SNOMED (\%)} & \textbf{Cog (\%)} \\
\midrule
Mistral    & MCoT & 72.5 & 40.0 & 60.0 \\
DeepSeek-R1-8B & CoT  & 58.3 & 30.0 & \cellcolor{green!20}70.0 \\
Gemma3 & MCoT & \cellcolor{green!20}85.0 & \cellcolor{green!20}50.0 & 50.0 \\
GPT-OSS & CoT  & 52.5 & 35.3 & 64.7 \\

\bottomrule
\end{tabular}

\vspace{2pt}
\footnotesize{Green highlights indicate highest effectiveness and dominant concept usage across models.}
\end{table}

\section{Linguistic Shift Thresholds via Otsu's Method}
\label{app:otsu}

\label{sec:otsu}
For each turn we define $\textbf{shift} = d_{\cos}(\text{MCoT}, \text{patient}) - d_{\cos}(\text{baseline}, \text{patient})$, the change in distance to the patient's utterance when MCoT replaces the baseline response. To surface dialogues with markedly larger shifts without an arbitrary cutoff, we apply Otsu's thresholding method~\cite{otsu1979threshold}, which automatically finds the split point that best separates a one-dimensional distribution into ``high'' and ``low'' regions. For each model and CBT principle, we compute per-transcript mean shift values and apply Otsu's method to separate high-shift transcripts, where MCoT meaningfully redirects the conversation, from low-shift ones (which stay closer to baseline). Flagging the high-shift transcripts (those above the threshold) identifies where guidance had the most effect, marking significant, therapeutically intended redirections for closer analysis.

The flagged counts confirm each model's default bias: \texttt{Mistral} and \texttt{Gemma3} redirect most consistently under \textit{(V)} (11/14 and, at a high 12.19\% threshold, 4/12), whereas \texttt{GPT-OSS} is the only model whose strongest redirection is \textit{(SQ)} (7/10). AP is rarely selected and never dominant, echoing the models' underuse of alternative perspectives.

\begin{table}[t]
\caption{
\textsc{Otsu-thresholded per-transcript Classification.} Mean
$|\text{shift}|$ (as \%). \textit{Threshold} is the Otsu cutoff;
\textit{Flagged} is the number of transcripts whose mean $|\text{shift}|$
exceeds the threshold, out of those in which the principle was selected.
Highlighted cells mark the highest-flagged protocol (V, SQ, AP) per model.}
\label{tab:otsu_shift}
\centering
\footnotesize
\setlength{\tabcolsep}{6pt}
\renewcommand{\arraystretch}{1.15}
\begin{tabular}{l c c c}
\toprule
\textbf{Model} & \textbf{Principle} & \textbf{Otsu-Threshold (\%)} & \textbf{Flagged} \\
\midrule
\multirow{4}{*}{Mistral}
  & Overall & 3.46  & 7/14 \\
  & \cellcolor{green!20}V & \cellcolor{green!20}0.95 & \cellcolor{green!20}11/14 \\
  & SQ      & 12.41 & 3/11 \\
  & AP      & 8.34  & 4/9 \\
\midrule
\multirow{4}{*}{Gemma3}
  & Overall & 4.13  & 6/14 \\
  & \cellcolor{green!20}V & \cellcolor{green!20}12.19 & \cellcolor{green!20}4/12 \\
  & SQ      & 7.93  & 1/14 \\
  & AP      & ---   & --- \\
\midrule
\multirow{4}{*}{GPT-OSS}
  & Overall & 4.56  & 4/14 \\
  & V       & 6.18  & 6/14 \\
  & \cellcolor{green!20}SQ & \cellcolor{green!20}4.37 & \cellcolor{green!20}7/10 \\
  & AP      & ---   & --- \\
\bottomrule
\end{tabular}
\end{table}

\bibliographystyle{IEEEtran}
\bibliography{references}

\end{document}